\documentclass[10pt,twocolumn,letterpaper]{article}

\usepackage{wacv}
\usepackage{times}
\usepackage{epsfig}
\usepackage{graphicx}
\usepackage{amsmath}
\usepackage{amssymb}
\usepackage{subfigure}
\usepackage{multirow}

\wacvfinalcopy %

\def\wacvPaperID{1002} %

\ifwacvfinal\fi
\begin{document}

\title{Human Annotations Improve GAN Performances}

\author{Juanyong Duan \\
National University of Singapore\\
{\tt\small j.duan@u.nus.edu}
\and
Sim Heng Ong \\
National University of Singapore\\
{\tt\small eleongsh@nus.edu.sg}
\and
Qi Zhao \\
University of Minnesota\\
{\tt\small qzhao@cs.umn.edu}
}

\maketitle
\ifwacvfinal\thispagestyle{empty}\fi

\begin{abstract}
Generative Adversarial Networks (GANs) have shown great success in many applications. In this work, we present a novel method that leverages human annotations to improve the quality of generated images. Unlike previous paradigms that directly ask annotators to distinguish between real and fake data in a straightforward way, we propose and annotate a set of carefully designed attributes that encode important image information at various levels, to understand the differences between fake and real images. Specifically, we have collected an annotated dataset that contains 600 fake images and 400 real images. These images are evaluated by 10 workers from the Amazon Mechanical Turk (AMT) based on eight carefully defined attributes. Statistical analyses have revealed different distributions of the proposed attributes between real and fake images. These attributes are shown to be useful in discriminating fake images from real ones, and deep neural networks are developed to automatically predict the attributes. We further utilize the information by integrating the attributes into GANs to generate better images. Experimental results evaluated by multiple metrics show performance improvement of the proposed model.
\end{abstract}

\section{Introduction}

Recently, generative adversarial networks (GANs) \cite{ian2014gan} have achieved impressive success in various applications \cite{zhu2017unpaired,zhang2017deep,van2016pixel,liu2016coupled,luc2016semantic}. A vanilla GAN contains a generator that maps a low-dimension latent code into the target space, for example, the image space. Instead of estimating the likelihood of the generated sample, it employs a discriminator to judge how difficult it is to discriminate from real samples. The generator and the discriminator are optimized jointly in an adversarial manner until equilibrium state has been reached. 

\begin{figure}[htbp]
	\centering
	\subfigure[Real images]{
		\label{real_sample}
		\includegraphics[scale=0.4]{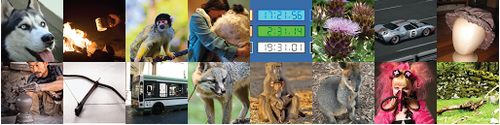}}
	\subfigure[Fake images]{
		\label{fake_sample}
		\includegraphics[scale=0.4]{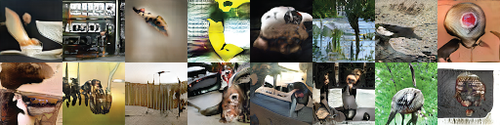}}
	\caption{Sample images from our dataset.}
	\label{sample}
\end{figure}

However, the training dynamics of GAN are usually unstable and the generator may output images that collapse to limited modes or with low quality. In this work, we aim to study whether human annotations will improve the quality of the generated images. To achieve this, we have collected human annotated data from the Amazon Mechanical Turk (AMT) on 1000 images, including 600 generated images (fake images), and for reference, 400 realistic images (real images). Figure \ref{sample} shows sample images from our dataset. We have defined eight attributes, namely, ``color", ``illuminance", ``object'', ``people'', ``scene'', ``texture'', ``realism'', and ``weirdness''. Each image has been annotated by 10 workers with a scale of 1 to 5 on the eight attributes. We have further analyzed the annotations and identified key features that contribute to image quality evaluation. Furthermore, we have also constructed deep neural networks to predict attributes of images to mimic human annotations. We find that integrating these attributes can improve the quality of generated images as evaluated by multiple metrics.

Our core insight is that we can leverage human annotations to encode image samples by treating them as prior knowledge to help the discriminator figure out fake images. Concretely, our contributions are twofold:
\begin{itemize}
\item We collect a new dataset of 1000 images with annotations from human to study the differences between real and fake images with a thorough analysis. In addition, we train the attribute net that mimics human subjects to annotate the attributes of new images.
\item We propose a new paradigm that includes the attribute net in the adversarial networks to improve the quality of generated images. Our paradigm can be applied to different GAN architectures and experiments have shown improvements in multiple metrics against baseline models.
\end{itemize}

\section{Related Work}\label{related_work}
\paragraph{Generated Adversarial Networks (GANs)}
GANs are a class of generative models that transform a known distribution (e.g. the normal distribution) to an unknown distribution (e.g. the image distribution). The two components, a generator and a discriminator, serve for different purposes. Informally, the generator is trained to fool the discriminator; thus we would like to improve the error rate of the discriminator during training, while the discriminator is trained to identify the generated samples from real samples. GANs have achieved impressive success in various applications, such as image generation \cite{radford2015unsupervised,zhao2016energy}, image editing \cite{zhu2016generative}, representation learning \cite{mathieu2016disentangling}, super-resolution \cite{ledig2016photo}, and domain transferring \cite{zhu2017unpaired,reed2016generative,isola2017image}.
\paragraph{Variants of GAN}
The vanilla GAN has many drawbacks, such as easily collapsing to a single point and generating nonsensical outputs. Radford et al. \cite{radford2015unsupervised} propose Deep Convolutional GANs (DCGANs) to produce more meaningful results. Although many attempts have been made to use convolutional nets to scale up GANs, such efforts were unsuccessful. They summarize several architecture guides for stable deep convolutional net-based GANs. Experiments on several benchmark datasets prove the effectiveness of the proposed guidelines. Another variant of vanilla GAN focuses on the training objective. Mao et al. \cite{mao2017least} propose a least square loss that mitigates the gradient vanishing problem when updating the generator. 

Inspired by theories in optimal transport, Arjovsky et al. \cite{arjovsky2017wasserstein} proposes the Wasserstein GAN which considers optimizing GAN as minimizing the Wasserstein distance between the distributions of real and fake images. The loss function requires the neural networks to be 1-Lipschitz. To achieve this, all the weights of the models are clipped within a range, but this method makes most of the weights fall on the boundary of the clipping interval. To make the weights distribute more naturally, Gulrajani et al. \cite{gulrajani2017improved} use a gradient penalty to regularize the 1-Lipschitz property on both the generator and the discriminator. Experiments have shown that Wasserstein GAN outperforms previous GAN variants and the gradient penalty performs better than weight clipping.

Recently, two large scale models, BigGAN \cite{brock2018large} and StyleGAN \cite{karras2019style} are proposed to synthesis images with high fidelity. BigGAN scales up traditional GAN with orthogonal regularization to the generator for finer control over the trade-offs between sample fidelity and variety. StyleGAN adopts style transfer methods by changing styles of latent codes to improve the control of strength of image features at different scales. However, both models require huge consumptions of computational power.

\section{Dataset}
In this section, we introduce the method used to generate the fake stimuli, including training details. We also describe the protocol for collecting data from AMT.
\subsection{Network Architecture}\label{wgangp-arch}
\begin{figure*}[htbp!]
	\centering
	\subfigure[Generator's architecture.]{
		\label{generator}
		\includegraphics[scale=0.28]{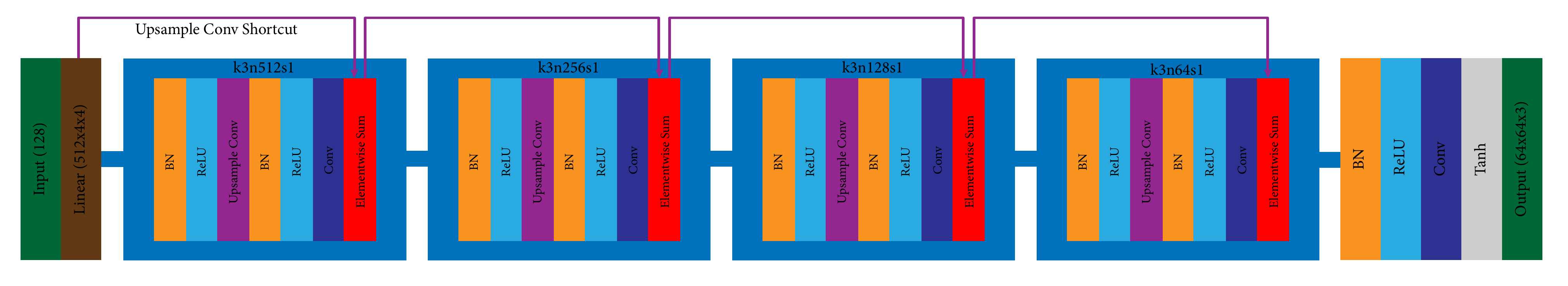}}
	\subfigure[Discriminator's architecture.]{
		\label{discriminator}
		\includegraphics[scale=0.28]{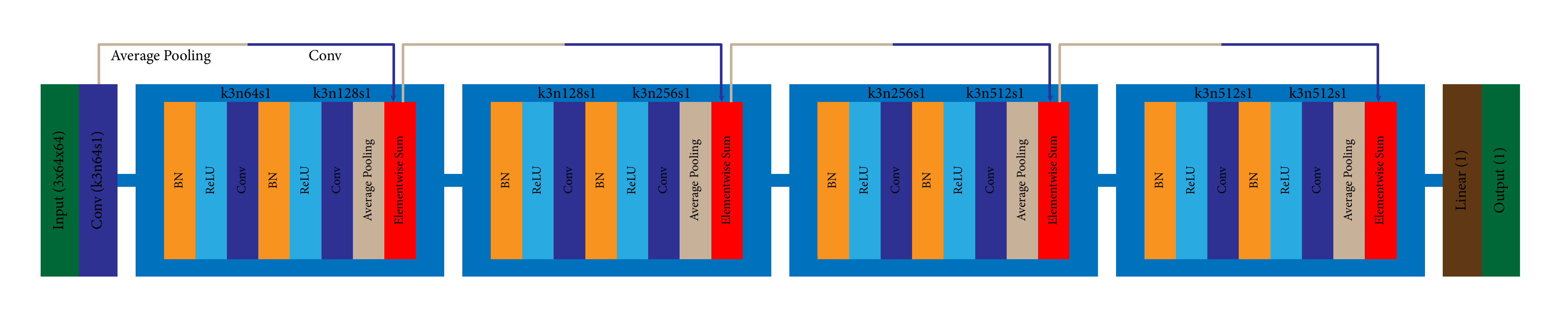}}
	\caption{Generator and discriminator architectures. \texttt{k3n512s1} means that the kernel size is 3, output dimension is 512 and stride is 1 for both upsample convolution layer and convolution layer. (a) Generator's architecture. (b) Discriminator's architecture.}
	\label{I}
\end{figure*}

The generator's architecture is shown in Figure \ref{generator}. We followed the implementation in \cite{gulrajani2017improved} to generate decent samples. The network uses 4 blocks of residual modules. Each residual module has an upsample convolution layer that uses a sub-pixel CNN \cite{shi2016real} for feature map upscaling. The shortcut path for each residual module is another upsample convolution layer that has the same output dimension as the one in the main path. At the end of the network, we use an additional layer of convolution and a $\tanh$ function to output the final image. The discriminator's architecture is shown in Figure \ref{discriminator}. The discriminator also has 4 residual modules. It uses average pooling to downscale the feature maps by a factor of 2 in the width and height dimensions. The shortcut path in the discriminator contains an average pooling layer and a convolution layer. The average pooling layer downscales feature maps by a factor of 2, while the convolution layer outputs feature maps that have the same dimension as the main path. At the end of the net, a linear transformation transforms the output to a single scalar that represents the probability that the sample came from the data distribution rather than the noise distribution.

The vanilla GAN may produce poor results due to unstable training, which may occur quite often \cite{metz2016unrolled,poole2016improved,salimans2016improved,arjovsky2017wasserstein}. Therefore, we use the Wasserstein GAN (WGAN) with gradient penalty paradigm \cite{gulrajani2017improved}. The objective function is defined by
\begin{align}
L = \mathbb{E}_{\tilde{\boldsymbol{x}}\sim \mathbb{P}_g}\left[D(\tilde{\boldsymbol{x}})\right]- \mathbb{E}_{\boldsymbol{x}\sim \mathbb{P}_r}\left[D(\boldsymbol{x})\right]\nonumber\\
+\lambda \mathbb{E}_{\hat{\boldsymbol{x}}\sim \mathbb{P}_{\hat{\boldsymbol{x}}}}\left[\left(\left\|\nabla_{\hat{\boldsymbol{x}}} D(\hat{\boldsymbol{x}})\right\|_2-1\right)^2 \right] \label{wgan_loss}
\end{align}
where $\tilde{\boldsymbol{x}}=G_{\boldsymbol{\theta}}(\boldsymbol{z})$ is drawn from the generator distribution $\mathbb{P}_g$, $\tilde{\boldsymbol{x}}$ is drawn from the real image distribution $\mathbb{P}_\text{data}$, and $\hat{\boldsymbol{x}}=\epsilon\boldsymbol{x}+(1-\epsilon)\tilde{\boldsymbol{x}}$ is a linear interpolation with a random number $\epsilon\sim U[0,1]$. $\lambda$ is the gradient penalty coefficient.

\subsection{Training Details}\label{wgangp-train}
The training procedure follows \cite{gulrajani2017improved}. We initialized the weights with He's normal initialization \cite{he2015delving}. We set the gradient penalty coefficient $\lambda$ to 10 and updated the discriminator once every 5 iterations. All the models were optimized with an Adam optimizer with $\beta_1=0$ and $\beta_2=0.9$. We set the initial learning rate $1\times10^{-4}$ and batch size 64. We trained the models for 200000 iterations. We sampled real images from the small ImageNet dataset with a resolution of 64 by 64 pixels \cite{van2016pixel,chrabaszcz2017downsampled}.
\subsection{Attributes to Annotate}
To accurately compare the quality of fake stimuli with real stimuli, we define a set of 8 attributes from different aspects. The attributes are listed in Table~\ref{table:labels}.

Pixel-level image attributes, such as color, intensity, and orientation, are low-level features for saliency detection and are biologically plausible \cite{itti1998model}. These features may be important to influence the quality of fake images. However, the stimuli are very small (64 by 64 pixels), so detailed information cannot be perceived easily. We only include color here. On the other hand, illuminance is an important attribute that differs from real images to computer generated images \cite{meyer1986experimental,mcnamara2005exploring}. We asked annotators to describe their feelings about the illuminance of the stimuli. 

Human attention tends to be drawn by objects relating to humans, such as faces \cite{kanwisher1997fusiform,cerf2009faces}, emotion \cite{adolphs2010does}, and crowds \cite{jiang2014saliency}. It is also easily attracted by moving objects \cite{kourtzi2000activation,winawer2008motion}. In addition, a key criteria is whether GANs can generate human recognizable objects in the image. Hence, we include object and human in the attribute list.

The classification of indoor/outdoor scenes is an important problem in computer vision \cite{payne2005indoor}. It has a well defined constraint that an outdoor scene is inside a man-made structure \cite{chen2016indoor}. The same semantic object may not be helpful to classify whether an image is indoor or outdoor, such as an indoor swimming pool and an outdoor swimming pool. In addition, some outdoor images are easier to generate, such as sky, ocean, and grassland. As a result, we include scenes in the attribute list to ask subjects to identify whether the image is perceived as indoor or as outdoor image.

In some previous image synthesis models, repeated patterns may be observed frequently in synthesized images \cite{portilla2000parametric}. It is possible for GANs to generate such patterns, so we define such features as texture and include it in our attribute list.

Realism is included as well. We also add weirdness in the set. Weirdness is defined as any unnatural features or objects, which might be common in generated images. Perception may be affected by the size of image \cite{chu2013size}. Real images may also contain objects that could be perceived strange if the image is small. 
\setlength{\tabcolsep}{4pt}
\begin{table*}[!htbp]
	\centering
	\begin{tabular}{ll}
		\hline\noalign{\smallskip}
		Attribute & Description \\
		\noalign{\smallskip}
		\hline
		\noalign{\smallskip}
		Color & colorfulness, pixel value distribution\\
		Illuminance &  light effect, shadows, brightness\\
		Object & objects in the image excluding humans, like car, animal, or furniture\\ 
		People & humans in the image\\
		Scene &  outdoor rather than indoor scene\\
		Texture & repeated pattern \\
		Realism & overall naturalness, real or computer generated \\
		Weirdness & any unnatural feature,such as strange objects \\
		\hline
	\end{tabular}
	\caption{\upshape List of attributes and descriptions}
	\label{table:labels}
\end{table*}

\setlength{\tabcolsep}{1.4pt}
\subsection{Stimuli and Data Collection}
\begin{figure}[!htbp]
	\centering
	\includegraphics[scale=0.25]{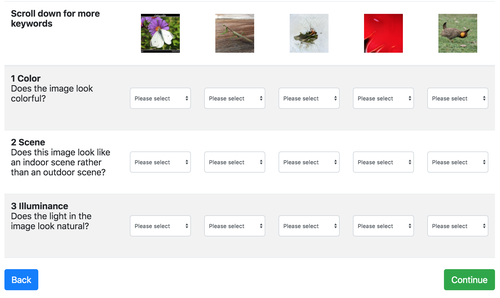}
	\caption{User interface in the AMT task.}
	\label{UI}
\end{figure}
The dataset contains 1000 images. We used the generator trained above to produce 600 fake images with a resolution of 64 by 64 pixels. Another 400 real images were randomly selected from the small ImageNet dataset. All the images from this dataset are downsampled from the original ImageNet dataset. Images from small ImageNet are similar to CIFAR-10 images \cite{torralba200880}, but have greater variety \cite{van2016pixel,chrabaszcz2017downsampled}. We requested workers from AMT to annotate the images. The user interface for the task is shown in Figure \ref{UI}.

Each worker was asked to annotate a total of 20 images in one assignment. Participants were asked to judge whether the keyword (attribute) best describes the stimuli. There are five choices, ``definitely yes" (5), ``probably yes" (4), ``not sure" (3), ``probably no" (2), and ``definitely no" (1). Each choice was then converted to the numerical score in the parentheses. To ensure the quality of annotation, we have set up two standards. Firstly, each participant must have an overall approval rate better than $95\%$. Secondly, at the end of the annotation, participants needed to annotate five extra images that were selected from the images they had annotated in current assignment. In addition, participants were not allowed to see their previous selections. If the score difference between two annotations is large (\textit{i.e.} $>1$), the current assignment would be rejected. For each image, we collected annotations from 10 participants which all met the requirements. We calculated the score vector by averaging the annotations from the 10 participants for each image.

\section{Data Analysis}
In this section, we summarize the statistical and factor analyses of the the annotated data.
\subsection{Statistical Analysis} 
\label{SA}
\subsubsection{Group means}
\begin{figure}[!hptb]
	\centering
	\includegraphics[scale=0.3]{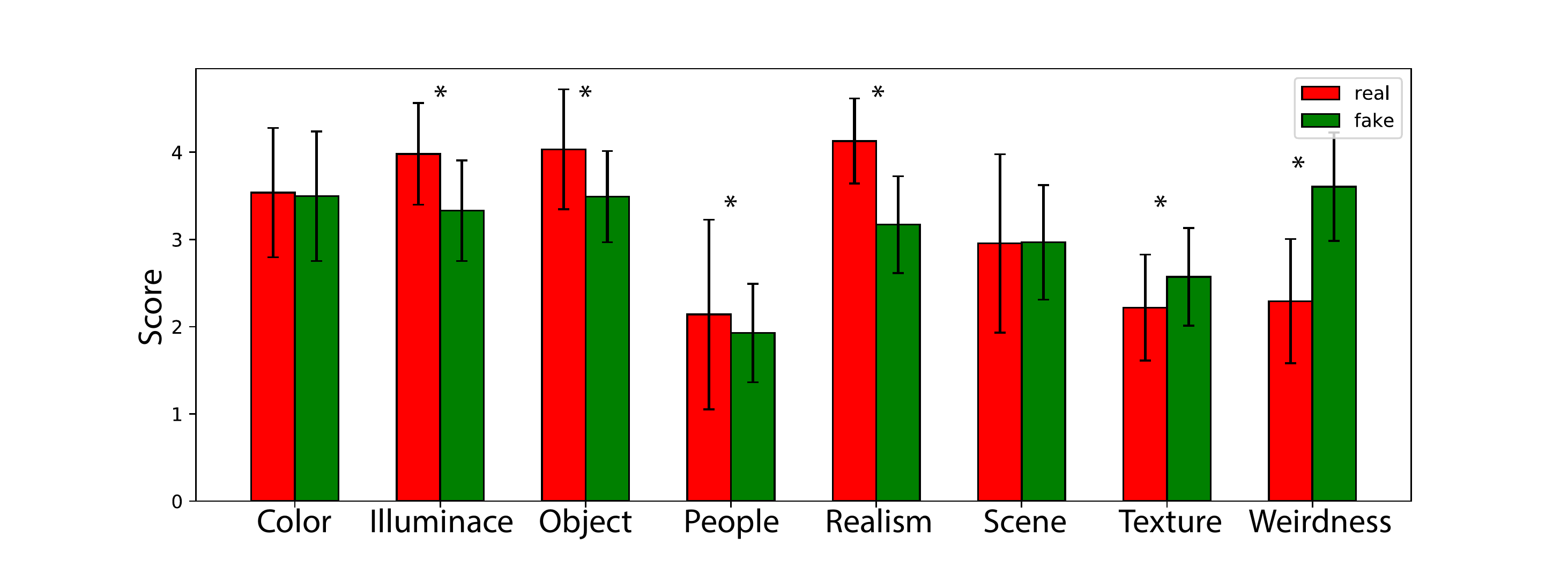}
	\caption{Bar plot of group means. `*' indicates statistically different ($p<0.01$).}
	\label{bp}
\end{figure}

We examined the sample mean of each attribute between real and fake images. We applied z-test on both groups of images. Figure \ref{bp} and Table \ref{table:stats} summarize the mean and standard deviation for each attribute.

Real images have higher scores in illuminance, object, people, and realism features, while fake images have higher scores in texture and weirdness features. There are no significant differences in color and scene features between real and fake images.

This observation shows that WGANs can generate colorful images that have similar color spectra as real images. However, WGANs are less capable of generating meaningful objects or humans. Fake images tend to be perceived as repeated patterns. Not surprisingly, fake images are rated more weird and less real.

\setlength{\tabcolsep}{4pt}
\begin{table}[hbtp]
	\centering
	\begin{tabular}{lcccc}
		\hline\noalign{\smallskip}
		Attribute & Real & Fake & $z$-value & $p$-value\\
		\noalign{\smallskip}
		\hline
		\noalign{\smallskip}
		Color&$3.54\pm0.74$&$3.50\pm0.74$&0.87&0.39\\
		Illuminance&$3.98\pm0.58$&$3.33\pm0.58$&17.38&$<0.01$\\
		Object&$4.03\pm0.69$&$3.50\pm0.52$&13.44&$<0.01$\\
		People&$2.14\pm1.09$&$1.93\pm0.56$&3.59&$<0.01$\\
		Realism&$4.13\pm0.49$&$3.17\pm0.55$&28.85&$<0.01$\\
		Scene&$2.95\pm1.02$&$2.97\pm0.66$&-0.21&0.83\\
		Texture&$2.22\pm0.61$&$2.57\pm0.56$&-9.28&$<0.01$\\
		Weirdness&$2.29\pm0.71$&$3.60\pm0.62$&-30.00&$<0.01$\\
		\hline\noalign{\smallskip}
	\end{tabular}
\caption{\upshape Summary of statistical results. Group means, standard deviations, $z$ values, and $p$ values are reported.}
\label{table:stats}
\end{table}
\setlength{\tabcolsep}{1.4pt}

\begin{figure*}[htbp!]
	\centering
	\subfigure[All images.]{
		\label{all_cor}
		\includegraphics[width=0.3\textwidth]{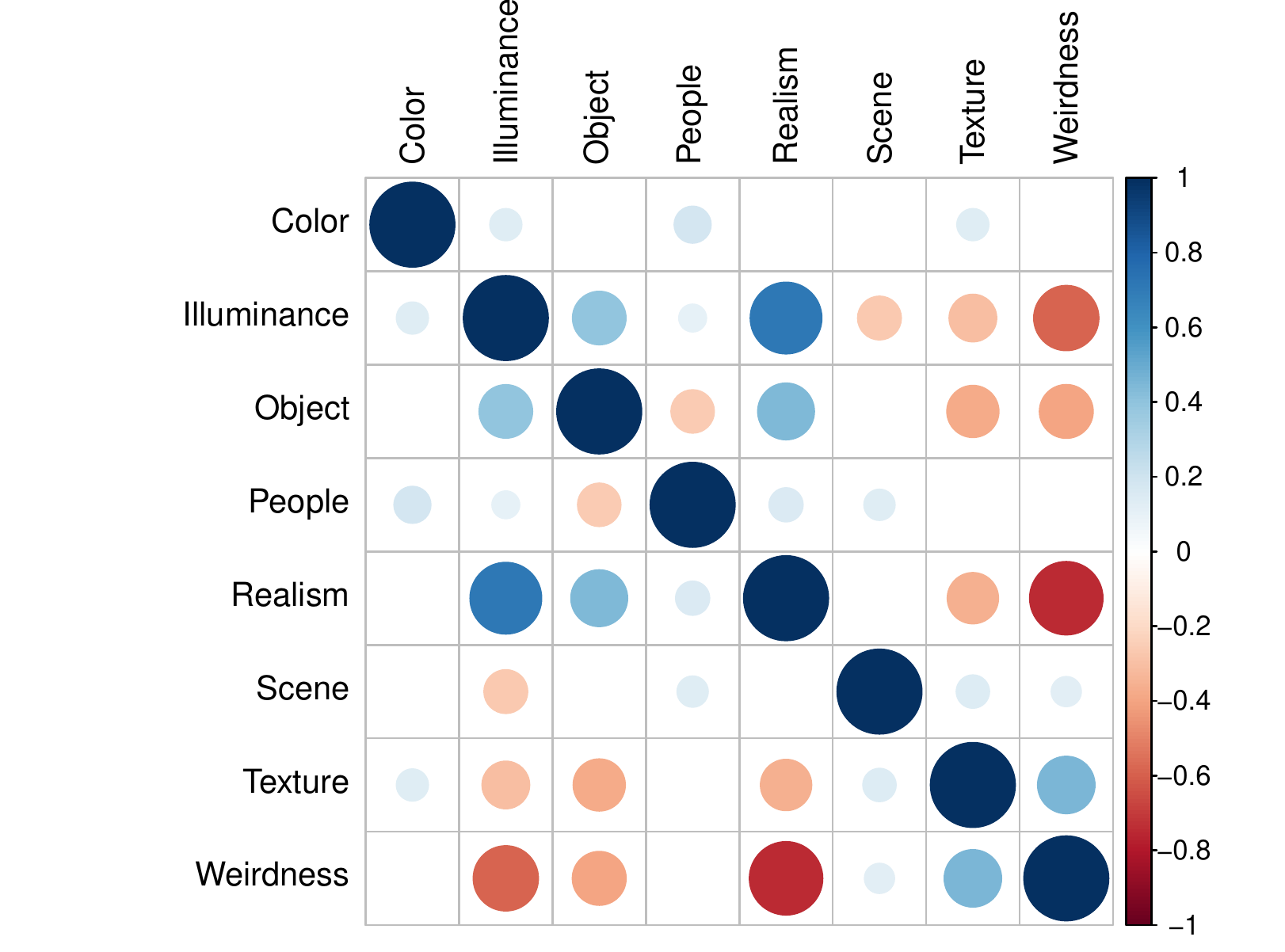}
	}
	\subfigure[Real images.]{
		\label{real_cor}
		\includegraphics[width=0.3\textwidth]{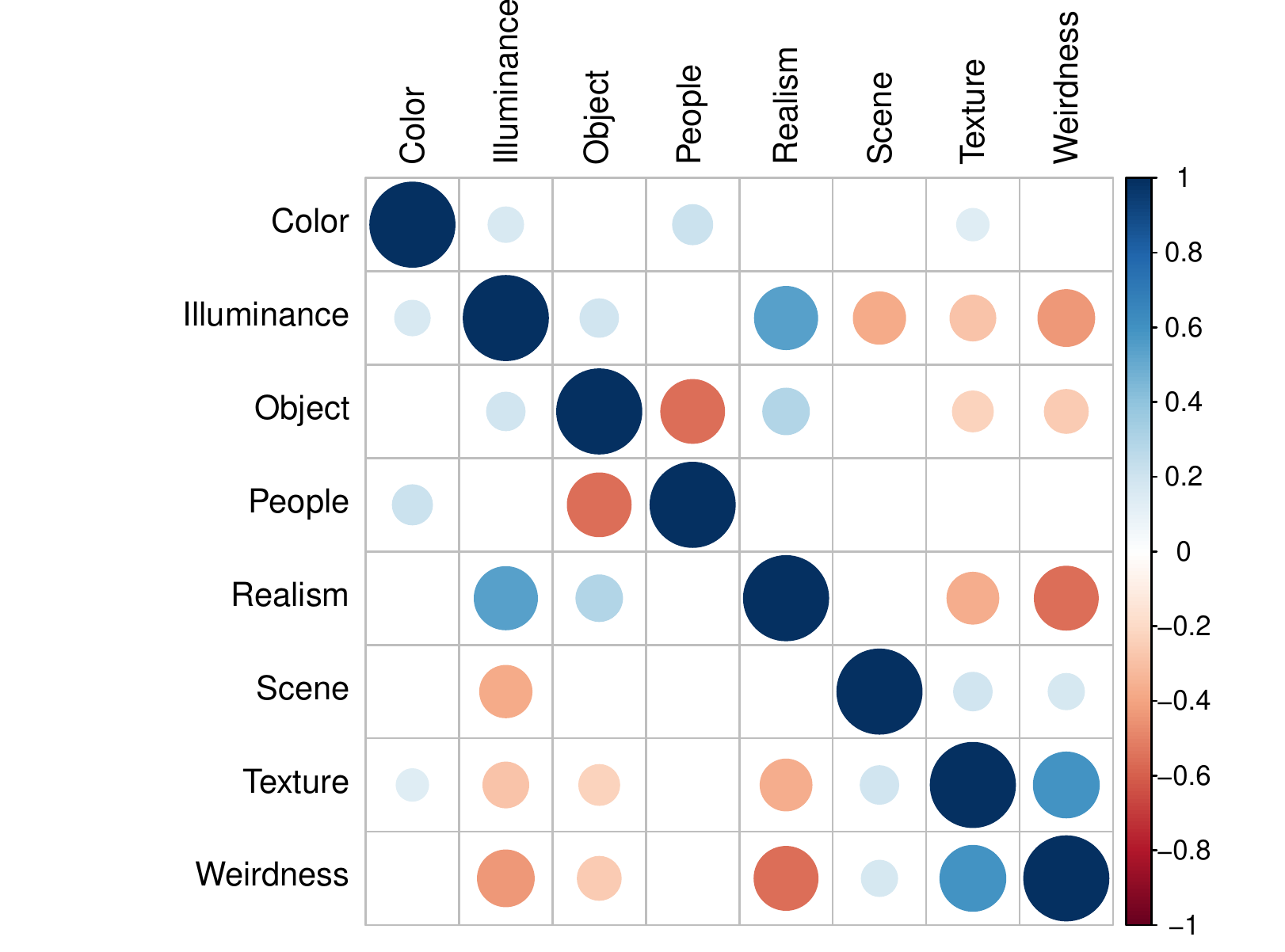}}
	\subfigure[Fake images.]{
		\label{fake_cor}
		\includegraphics[width=0.3\textwidth]{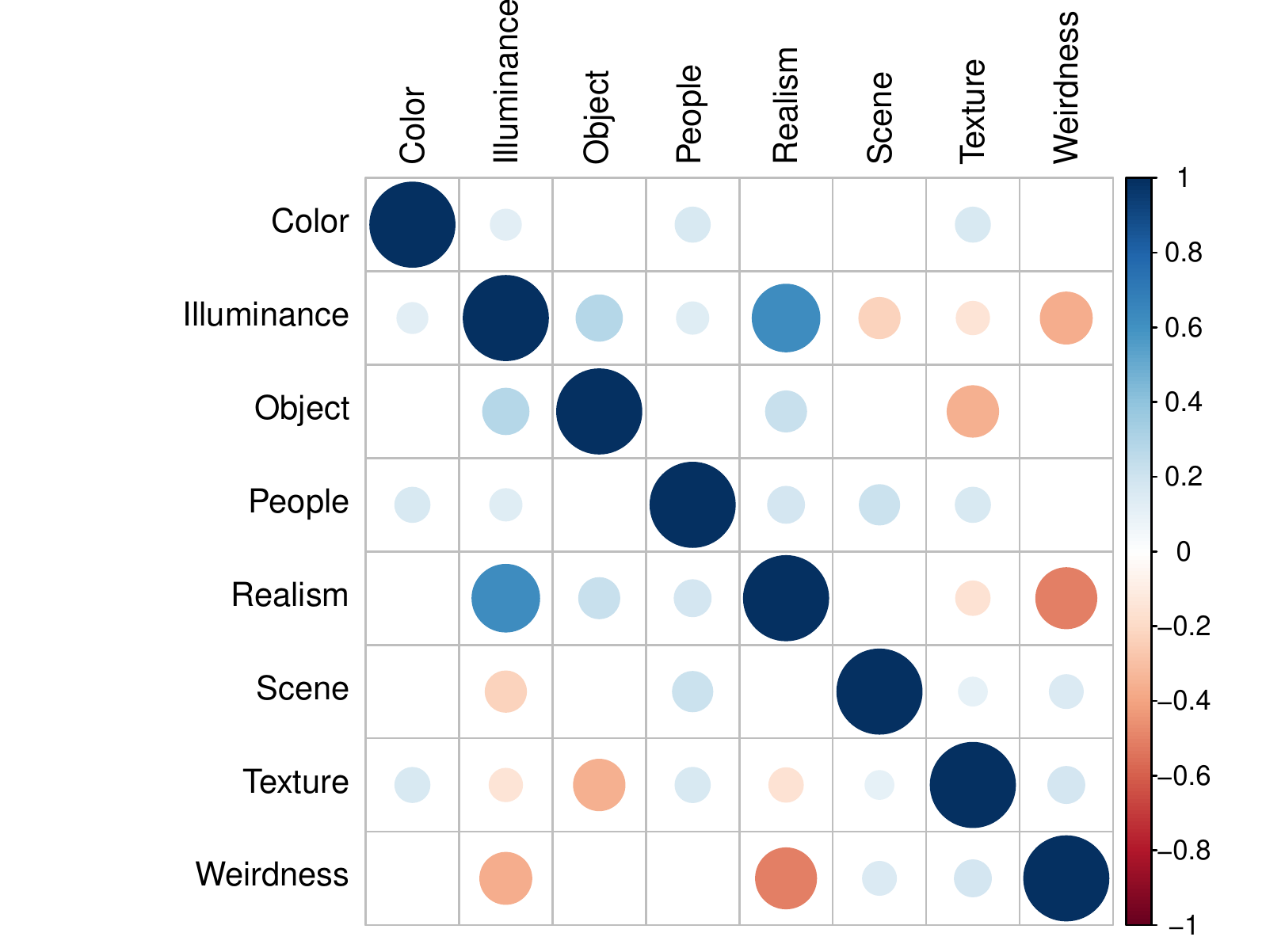}}
	\caption{Correlation matrices between attributes.}
	\label{correlation}
\end{figure*} 
\subsubsection{Correlation}We analyzed relationships between attributes by computing Pearson correlation coefficients. We first computed correlation with all the images, and then separately for real images and fake images. Figure \ref{correlation} shows the results. A blank cell indicates insignificant correlation ($p>0.01$). We observe that illuminance is highly correlated with realism ($r=0.71$). Note the correlation between illuminance and realism within each group. These two attributes are also moderately correlated ($r=0.55$ for real images, $r=0.63$ for fake images.) This shows that illuminance is an important factor for realism. This result is consistent with previous findings \cite{meyer1986experimental,mcnamara2005exploring,fan2012real,fan2017image}. We can also observe that object and realism are also correlated ($r=0.44$ for all images). The result accords with previous findings that images with more objects are perceived to be more realistic \cite{fan2017image,lalonde2007using,choi2009investigation}. We notice that for real images, object and human are negatively correlated ($r=-0.57$). This is because the stimuli are very small, so a single image is not likely have both objects and human. But for fake images, the correlation between them is insignificant. This indicates that WGANs are not likely to generate meaningful objects or humans. As expected, realism is negatively correlated with weirdness ($r=-0.74$).

\subsection{Factor Analysis}
Factor analysis (FA) is a statistical method that describes observed variables by latent, unobserved factors. Factor analysis is similar to principal component analysis (PCA) in that both are feature dimension reduction methods. However, components in PCA must be orthogonal to maximize the total variance, but the factors in FA are not necessarily orthogonal so that they can correlate with each other. We applied exploratory factor analysis (EFA) followed by confirmatory factor analysis (CFA) on the whole dataset. EFA identifies latent factors as linear combination of observed variables, while CFA tests how well the model fits the data. Attributes with poor fits, or loadings, are eliminated. 

\begin{figure}
	\centering
	\includegraphics[scale=0.4]{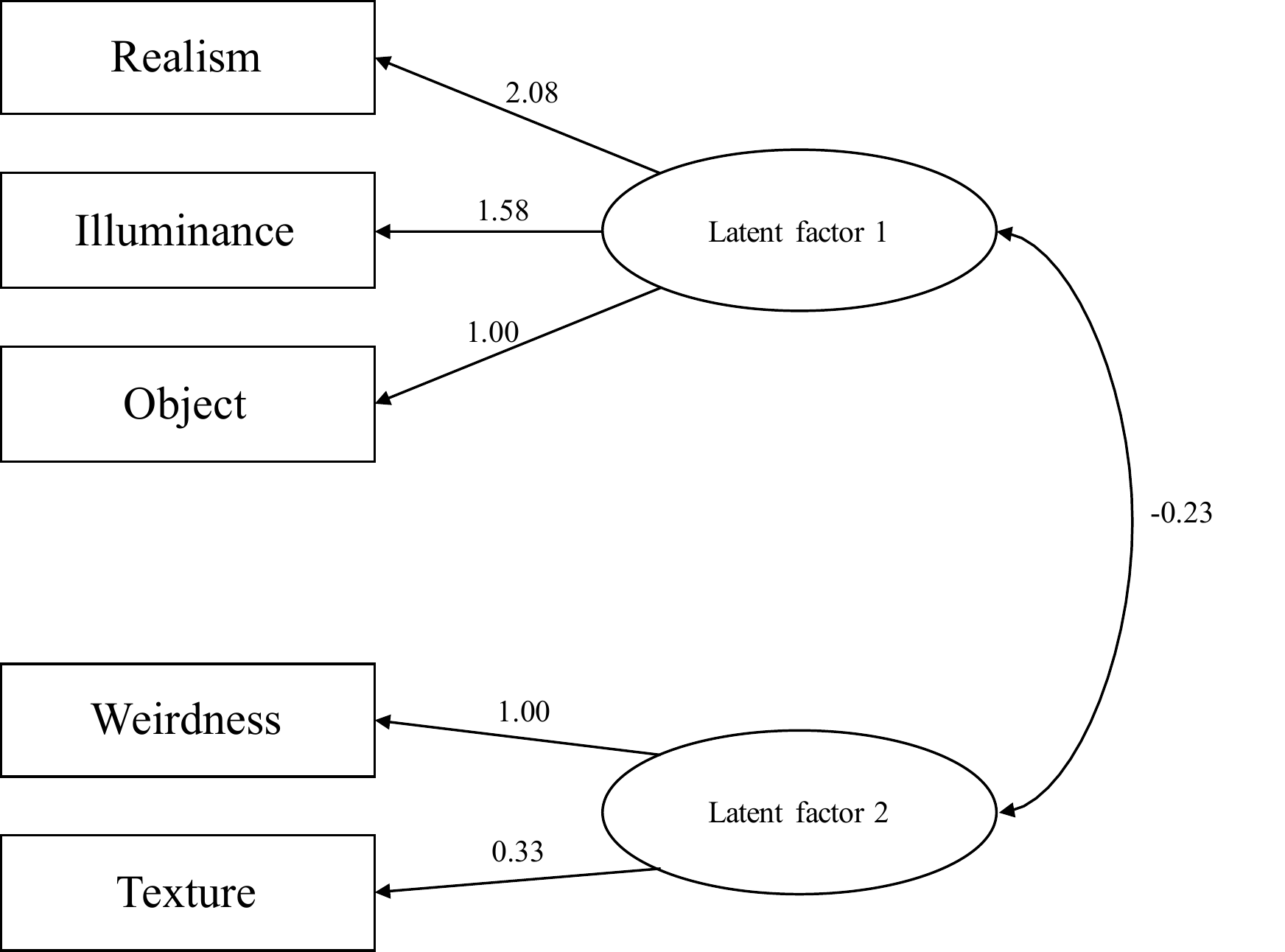}
	\caption{Results for factor analysis. The numbers are parameters estimated in the model.}
	\label{fa}
\end{figure}

The model parameters and structure are presented in Figure \ref{fa}. To estimate the fit of the model, two common indices are used. The first one is the Comparative Fit Index (CFI), which compares a chi-square for the fit of a target model to the chi-square for the fit of an independence model, \textit{i.e}., one in which the variables are uncorrelated. Higher CFIs indicate better model fit. Values
that approach 0.90 indicate acceptable fit \cite{fan2017image,kline1999principles}. Another model fit metric is Root Mean Square Error of Approximation (RMSEA), which estimates the amount of error of approximation per model degree of freedom and takes sample size into account. Smaller RMSEA values suggest better model fit. A value of 0.10 or less is indicative of acceptable model fit \cite{fan2017image,kline1999principles}. Our CFA model has acceptable fit, CFI = 0.97, RMSEA = 0.117.

As indicated in Figure \ref{fa}, we identified 2 latent factors. ``Latent factor 1" is measured by realism, illuminance, and object. ``Latent factor 2" is measured by weirdness and texture. The result is consistent with the analysis in \ref{SA}. Realism, illuminance, and object are important factors for realism, while weirdness and texture are characteristics for fake images. Hence, ``latent factor 1" is associated with reality while ``latent factor 2" is associated with fakeness.

\section{Improving GAN with Attributes}\label{improve_gan}
In this section, we show that the annotated attributes can be used to improve the quality of generated images. We first train an attribute net to mimic human subjects to annotate the attributes of images and then describe the structure of our model and explain how it utilizes these attributes. Quantitative results show that our model outperforms baseline models. We also provide some qualitative examples and conclude with a brief discussion.

\subsection{Models}
\begin{figure*}[htbp!]
	\centering
	\includegraphics[scale=0.3]{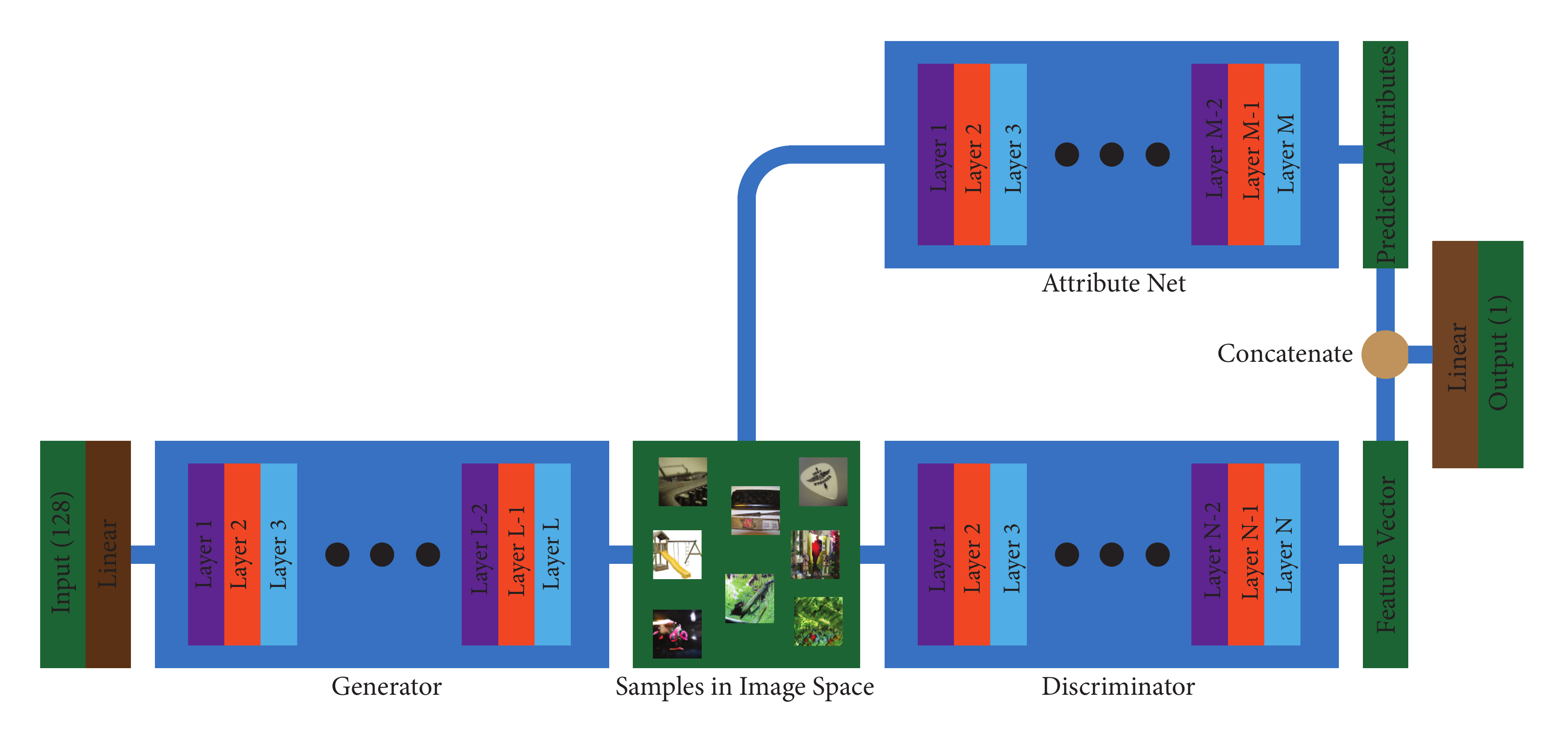}
	\caption{Proposed model. The input is a random vector of dimension 128. The generator takes it as input and output a sample in the image space. Then the sample is fed into the attribute net and the discriminator simultaneously, whose outputs are concatenated together. The a linear layer computes the final scalar output which is used to compute the loss. The detailed architectures of the generator, the discriminator, and the attribute nets may vary and are discussed in Section \ref{improve_gan}.}
	\label{fig:net}
\end{figure*}
Our model is shown in Figure \ref{fig:net}. It consists of a generator, a discriminator, and an attribute net. The generator accepts a random sample $\mathbf{z}$ drawn from a prior, e.g. normal distribution, and outputs an image sample, which then fed into the discriminator and the attribute net simultaneously. The attribute net takes an image as input and outputs the attributes. The outputs of the discriminator and the attribute net are concatenated together, which then feed into a fully connected layer to compute the final output. All the components in our model can be implemented in different ways.

\paragraph{Attribute Net}We implement the attribute net with three models, VGG-16, ResNet50, and DenseNet169. In addition, we also conduct an experiment that substitutes the attribute net with random noise. The purpose is to check the effectiveness of different network architectures. We impose random noise to prove that only semantic vectors can improve the quality of generated images.

We train all the attribute nets on the annotated data. The dataset is randomly split into training and validation sets, each containing 500 images. All the images are resized to $224\times224$ pixels and normalized. The loss function is the mean squared error between predicted values and annotated values. We train the model for maximum 300 epochs. Training is stopped either the maximum epoch is reached or the loss plateaus on the validation set. All the models are optimized with the stochastic gradient descent method with a mini batch size 16.

\paragraph{Adversarial Net}We test the attribute net on three GAN variants, WGAN, DCGAN, and LSGAN. For a fair comparison, we use the original loss function and network architecture for each GAN.

\subsection{Datasets and Evaluation Metrics}
As our annotations are obtained on the down-sampled ImageNet dataset, we first evaluate our model on it. We also evaluate our model on the CIFAR-10 dataset, which consists of 60000 tiny images with a resolution of $32\times 32$. This dataset is widely used for GAN studies. Although we did not collect annotations from CIFAR-10, we wish to evaluate whether the annotated information is transferable.

We adopt three evaluation metrics that are commonly used in previous works, namely inception score \cite{salimans2016improved}, mode score \cite{che2016mode}, and the Fr\'{e}chet Inception Distance (FID) \cite{heusel2017gans}. The Inception Score evaluates the KL divergence between th e conditional label distribution computed by the Inception model pretrained on the ImageNet dataset and the distribution of category labels. The Mode Score is an improved version of the inception score. It has an additional term which computes the KL divergence between the marginal label distribution from generated samples and the data label distribution. Finally, the Fr\'{e}chet Inception Distance (FID) is the distance between the two Gaussian random variables $\phi(\mathbb{P}_r)$ and $\phi(\mathbb{P}_g)$, where $\phi$ is a predefined feature function. Let $\mu_r$ and $\mu_g$ be the empirical means, and $\mathbf{C}_r$ and $\mathbf{C}_g$ be the empirical covariance of $\phi(\mathbb{P}_r)$ and $\phi(\mathbb{P}_g)$ respectively. Then the Fr\'{e}chet distance is defined as
\[
\text{FID}(\mathbb{P}_g, \mathbb{P}_r)=\|\mu_r-\mu_g\|+\text{Tr}(\mathbf{C}_r+\mathbf{C}_g-2(\mathbf{C}_r\mathbf{C}_g)^{1/2}).
\]

\subsection{Quantitative Results}
\setlength{\tabcolsep}{4pt}
\begin{table*}[htbp!]\small 
	\centering
	\begin{tabular}{lccc}
		\hline\noalign{\smallskip}
		GAN Type & Inception Score & Mode Score & FID \\\hline
		WGAN-WC \cite{arjovsky2017wasserstein} & $7.14\pm0.11$ & $5.07\pm0.07$ & $0.524\pm0.002$ \\
		WGAN-GP \cite{gulrajani2017improved} & $9.91\pm0.11$ & $7.86\pm0.11$ & $0.463\pm0.004$ \\
		LSGAN \cite{mao2017least} & $7.80\pm0.12$ & $5.69\pm0.09$ & $0.510\pm0.004$ \\
		DCGAN \cite{radford2015unsupervised} & $7.50\pm0.08$ & $5.62\pm0.09$ & $0.500\pm0.004$ \\
		CTGAN (as reported in \cite{wei2018improving}) & $10.27\pm0.15$ & - & - \\\hline
		\textbf{WGAN+VGG} & $\mathbf{11.11\pm0.15}$ & $\mathbf{9.06\pm0.13}$ & $\mathbf{0.450\pm0.003}$\\
		\textbf{WGAN+ResNet} & $10.47\pm0.16$ & $8.74\pm0.15$ & $0.477\pm0.003$\\
		\textbf{WGAN+DenseNet} & $10.17\pm0.11$ & $8.26\pm0.17$ & $0.472\pm0.003$\\
		\textbf{WGAN+Random Noise} & $7.07\pm0.16$ & $5.42\pm0.13$ & $0.505\pm0.005$\\
		\hline
		\textbf{DCGAN+VGG} & $8.28\pm0.10$ & $7.13\pm0.11$ & $0.495\pm0.003$\\
		\textbf{LSGAN+VGG} & $8.12\pm0.14$ & $7.26\pm0.14$ & $0.488\pm0.002$\\
		\hline
	\end{tabular}
\caption{\upshape Quantitative results of unsupervised training on ImageNet. Best results are shown in bold. For inception score and mode score, higher score represents better quality. Lower FID indicates the fake distribution is closer to the real distribution.}
\label{table:perfs}
\end{table*}
\setlength{\tabcolsep}{1.4pt}
Quantitative results for models trained on ImageNet are summarized in Table \ref{table:perfs}.

\subsubsection{Training with Different Attribute Nets} 
\begin{figure}
	\centering
	\includegraphics[scale=0.4]{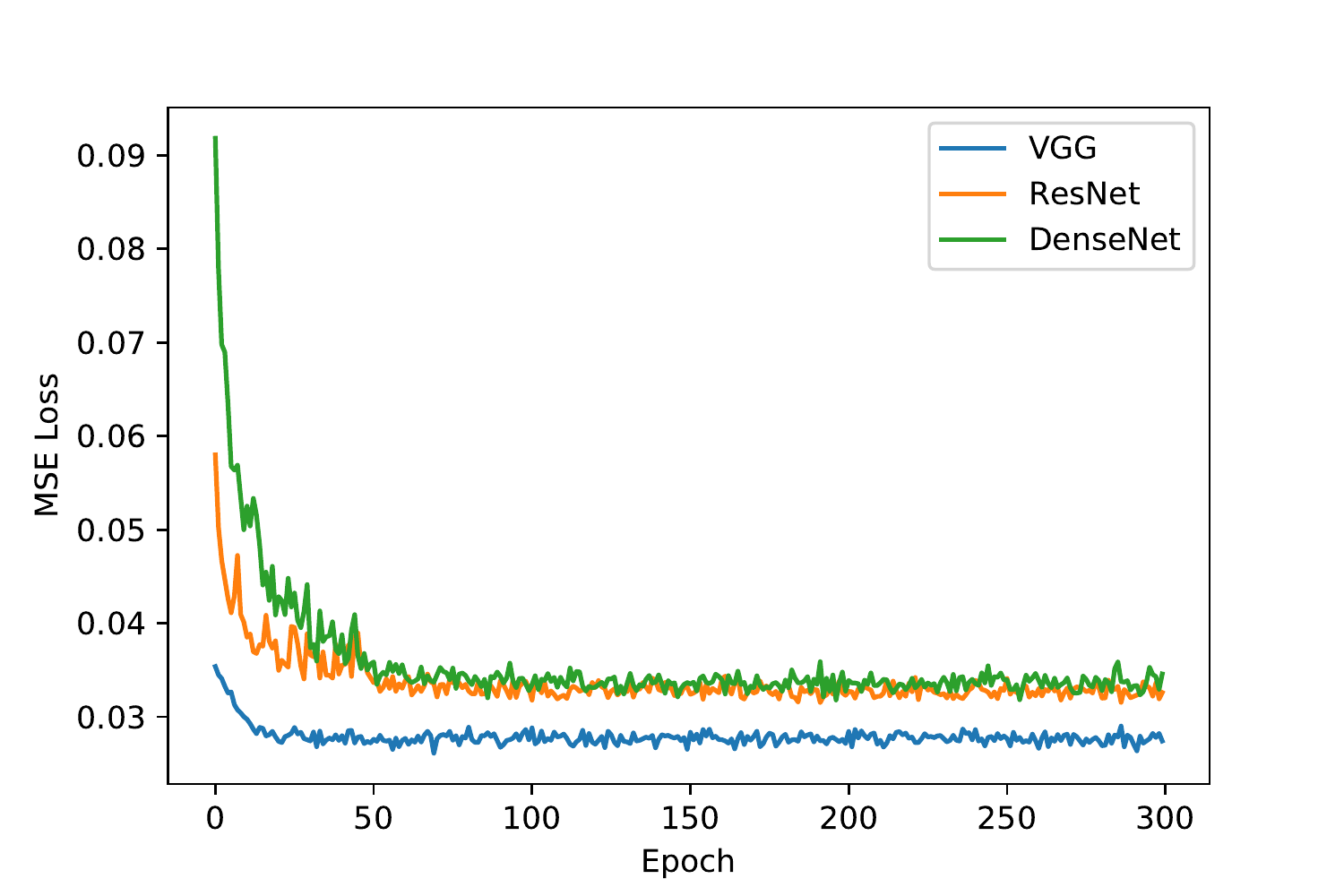}
	\caption{MSE Loss Curve of Attribute Net. VGG predicts the attributes more accurate than other two models.}\label{attribute_loss}
\end{figure}
We observe that VGG achieves the best performance among different architectures of the attribute net. VGG learns feature representations in an hierarchical way, meaning higher layers learn an ensemble of features from lower layers. ResNet and DenseNet are designed to learn residuals, making successive layers refine previous layers. This paradigm of learning may quickly learn representations for object classifications, but may perform poorly in transfer learning or fine tuning tasks. In other tasks like style transfer \cite{li2018closedform,huang2017arbitrary,gatys2015neural}, VGG is more popular than ResNet for extracting features because VGG requires fewer parameter tuning tricks and converges faster than ResNet and DenseNet. In our case, we are finetuning the pretrained model on a small dataset, therefore VGG may perform better than other two models. Figure \ref{attribute_loss} shows the RMSE loss when each model reaches stopping time. VGG16 has least RMSE loss, which indicates it predicts more accurate attributes.

\subsubsection{Training with Different GAN Variants}
We examine whether our proposed attributes are useful for different types of GANs. The results show that for three types of GANs, WGAN, DCGAN, and LSGAN, integrating attributes will lead to better performance. This is possibly because the proposed attributes represent higher level semantics that might not be learned directly from images. Additional information might make the discriminator discriminate fake samples easier. 

To ensure that only vectors with semantic means could improve the performance of GANs, we assign a random vector drawn from normal distribution to each image. As shown in Table \ref{table:perfs}, the inception score drops significantly for this case.

This phenomenon indicates that only when the input vector has some semantic means can the discriminator performs better. A random vector has little information about the input sample and thus may interfere the prediction of the discriminator, which causes the drops in evaluation metrics.
\subsubsection{Training on Different Datasets}
To examine whether the attributes are generalizable to other datasets, we train the model on CIFAR-10. However, as shown in Table \ref{cifar-result}, the inception score is lower than the WGAN-GP model, which indicates that the attributes may distribute inconsistently among different datasets. 

One possible reason is that image resolutions for two datasets are different. Therefore, the attribute net failed to compute the correct attribute scores for real images from CIFAR-10 and fake images generated by GAN. Consequently, the attribute scores may interfere the prediction of the discriminator, similar as a random vector.

\begin{table}[htbp!]
	\centering
	\begin{tabular}{cc}
		\hline
		Model & Inception Score \\\hline
		WGAN-GP & $7.86\pm0.07$ \\
		WGAN+VGG & $6.77\pm0.07$\\\hline
	\end{tabular}
	\caption{CIFAR-10 Results}\label{cifar-result}
\end{table}

\subsection{Qualitative Results}
We show some images generated by the WGAN+VGG combination (Figure \ref{fig:our_sample}). More samples are shown in the supplemental materials. We analyze the generated images qualitatively from three aspects, pixels, diversity, and reality.

\paragraph{Pixels} We examine pixel values by two factors, color and sharpness. We observe that the color looks natural and diverse in all images. The color distribution is consistent with natural images. Sharpness indicates that the difference between adjacent pixels is large. A sharper image looks less blurry and edges or boundaries can be figured out more easily, so images containing recognizable objects are usually sharp. As we can see from the sample images, almost all of them look sharp, which means that our model captures this feature successfully.

\paragraph{Diversity} A common failure of GANs is mode collapse, meaning that the same image is generated for different latent vectors. Hence, diversity is an important factor to evaluate the performance of GANs. From the generated samples we observe that they are quite diverse and we can hardly find same images.

\paragraph{Reality} Reality indicates whether the image contains recognizable objects. Unfortunately, we find that many images contain meaningless color blotches. But we can also find a few images have distorted dogs and cats, like the second and the seventh images in the first row of Figure \ref{fig:our_sample}.

\begin{figure}[htbp]
	\centering
	\includegraphics[scale=0.3]{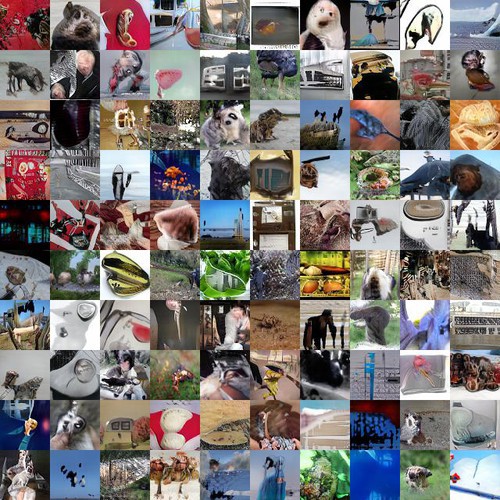}
	\caption{Qualitative Results.}
	\label{fig:our_sample}
\end{figure} 

\section{Conclusion and Future Works}
In this project, we built a new dataset of annotated images to characterize generated images. Comprehensive analyses show that real images contain more semantic objects, have better illuminance, and are perceived more real than fake images. Fake images tend to be perceived as being more weird and more like repeated patterns. Further, a DNN is trained to predict attributes automatically. We integrate the trained attribute net into the discriminator of GAN to improve its performance. For future studies, a larger dataset could be built with more structured attributes for a more comprehensive study. Moreover, annotated attributes may be used for conditioned training or disentangled feature learning.

{\small
\bibliographystyle{ieee}
\bibliography{gan}

\begin{thebibliography}{10}\itemsep=-1pt

\bibitem{adolphs2010does}
R.~Adolphs.
\newblock What does the amygdala contribute to social cognition?
\newblock {\em Annals of the New York Academy of Sciences}, 1191(1):42--61,
  2010.

\bibitem{arjovsky2017wasserstein}
M.~Arjovsky, S.~Chintala, and L.~Bottou.
\newblock Wasserstein gan.
\newblock {\em arXiv preprint arXiv:1701.07875}, 2017.

\bibitem{brock2018large}
A.~Brock, J.~Donahue, and K.~Simonyan.
\newblock Large scale gan training for high fidelity natural image synthesis.
\newblock {\em arXiv preprint arXiv:1809.11096}, 2018.

\bibitem{cerf2009faces}
M.~Cerf, E.~P. Frady, and C.~Koch.
\newblock Faces and text attract gaze independent of the task: Experimental
  data and computer model.
\newblock {\em Journal of vision}, 9(12):10--10, 2009.

\bibitem{che2016mode}
T.~Che, Y.~Li, A.~P. Jacob, Y.~Bengio, and W.~Li.
\newblock Mode regularized generative adversarial networks.
\newblock {\em arXiv preprint arXiv:1612.02136}, 2016.

\bibitem{chen2016indoor}
C.~Chen, Y.~Ren, and C.-C.~J. Kuo.
\newblock Indoor/outdoor classification with multiple experts.
\newblock In {\em Big Visual Data Analysis}, pages 23--63. Springer, 2016.

\bibitem{choi2009investigation}
S.~Y. Choi, M.~Luo, M.~Pointer, and P.~Rhodes.
\newblock Investigation of large display color image appearance--iii: Modeling
  image naturalness.
\newblock {\em Journal of Imaging Science and Technology}, 53(3):31104--1,
  2009.

\bibitem{chrabaszcz2017downsampled}
P.~Chrabaszcz, I.~Loshchilov, and F.~Hutter.
\newblock A downsampled variant of imagenet as an alternative to the cifar
  datasets.
\newblock {\em arXiv preprint arXiv:1707.08819}, 2017.

\bibitem{chu2013size}
W.-T. Chu, Y.-K. Chen, and K.-T. Chen.
\newblock Size does matter: How image size affects aesthetic perception?
\newblock In {\em Proceedings of the 21st ACM international conference on
  Multimedia}, pages 53--62. ACM, 2013.

\bibitem{fan2012real}
S.~Fan, T.-T. Ng, J.~S. Herberg, B.~L. Koenig, and S.~Xin.
\newblock Real or fake?: human judgments about photographs and
  computer-generated images of faces.
\newblock In {\em SIGGRAPH Asia 2012 technical briefs}, page~17. ACM, 2012.

\bibitem{fan2017image}
S.~Fan, T.-T. Ng, B.~L. Koenig, J.~S. Herberg, M.~Jiang, Z.~Shen, and Q.~Zhao.
\newblock Image visual realism: From human perception to machine computation.
\newblock {\em IEEE transactions on pattern analysis and machine intelligence},
  2017.

\bibitem{gatys2015neural}
L.~A. Gatys, A.~S. Ecker, and M.~Bethge.
\newblock A neural algorithm of artistic style, 2015.

\bibitem{ian2014gan}
I.~J. Goodfellow, J.~Pouget-Abadie, M.~Mirza, B.~Xu, D.~Warde-Farley, S.~Ozair,
  A.~Courville, and Y.~Bengio.
\newblock Generative adversarial nets.
\newblock In {\em NIPS}, 2014.

\bibitem{gulrajani2017improved}
I.~Gulrajani, F.~Ahmed, M.~Arjovsky, V.~Dumoulin, and A.~C. Courville.
\newblock Improved training of wasserstein gans.
\newblock In {\em Advances in Neural Information Processing Systems}, pages
  5769--5779, 2017.

\bibitem{he2015delving}
K.~He, X.~Zhang, S.~Ren, and J.~Sun.
\newblock Delving deep into rectifiers: Surpassing human-level performance on
  imagenet classification.
\newblock In {\em Proceedings of the IEEE international conference on computer
  vision}, pages 1026--1034, 2015.

\bibitem{heusel2017gans}
M.~Heusel, H.~Ramsauer, T.~Unterthiner, B.~Nessler, and S.~Hochreiter.
\newblock Gans trained by a two time-scale update rule converge to a local nash
  equilibrium.
\newblock In {\em Advances in Neural Information Processing Systems}, pages
  6626--6637, 2017.

\bibitem{huang2017arbitrary}
X.~Huang and S.~Belongie.
\newblock Arbitrary style transfer in real-time with adaptive instance
  normalization.
\newblock {\em 2017 IEEE International Conference on Computer Vision (ICCV)},
  Oct 2017.

\bibitem{isola2017image}
P.~Isola, J.-Y. Zhu, T.~Zhou, and A.~A. Efros.
\newblock Image-to-image translation with conditional adversarial networks.
\newblock {\em arXiv preprint}, 2017.

\bibitem{itti1998model}
L.~Itti, C.~Koch, and E.~Niebur.
\newblock A model of saliency-based visual attention for rapid scene analysis.
\newblock {\em IEEE Transactions on pattern analysis and machine intelligence},
  20(11):1254--1259, 1998.

\bibitem{jiang2014saliency}
M.~Jiang, J.~Xu, and Q.~Zhao.
\newblock Saliency in crowd.
\newblock In {\em European Conference on Computer Vision}, pages 17--32.
  Springer, 2014.

\bibitem{kanwisher1997fusiform}
N.~Kanwisher, J.~McDermott, and M.~M. Chun.
\newblock The fusiform face area: a module in human extrastriate cortex
  specialized for face perception.
\newblock {\em Journal of neuroscience}, 17(11):4302--4311, 1997.

\bibitem{karras2019style}
T.~Karras, S.~Laine, and T.~Aila.
\newblock A style-based generator architecture for generative adversarial
  networks.
\newblock In {\em Proceedings of the IEEE Conference on Computer Vision and
  Pattern Recognition}, pages 4401--4410, 2019.

\bibitem{kline1999principles}
R.~B. Kline and D.~A. Santor.
\newblock Principles \& practice of structural equation modelling.
\newblock {\em Canadian Psychology}, 40(4):381, 1999.

\bibitem{kourtzi2000activation}
Z.~Kourtzi and N.~Kanwisher.
\newblock Activation in human mt/mst by static images with implied motion.
\newblock {\em Journal of cognitive neuroscience}, 12(1):48--55, 2000.

\bibitem{lalonde2007using}
J.-F. Lalonde and A.~A. Efros.
\newblock Using color compatibility for assessing image realism.
\newblock In {\em Computer Vision, 2007. ICCV 2007. IEEE 11th International
  Conference on}, pages 1--8. IEEE, 2007.

\bibitem{ledig2016photo}
C.~Ledig, L.~Theis, F.~Husz{\'a}r, J.~Caballero, A.~Cunningham, A.~Acosta,
  A.~Aitken, A.~Tejani, J.~Totz, Z.~Wang, et~al.
\newblock Photo-realistic single image super-resolution using a generative
  adversarial network.
\newblock {\em arXiv preprint}, 2016.

\bibitem{li2018closedform}
Y.~Li, M.-Y. Liu, X.~Li, M.-H. Yang, and J.~Kautz.
\newblock A closed-form solution to photorealistic image stylization.
\newblock {\em Lecture Notes in Computer Science}, page 468–483, 2018.

\bibitem{liu2016coupled}
M.-Y. Liu and O.~Tuzel.
\newblock Coupled generative adversarial networks.
\newblock In {\em Advances in neural information processing systems}, pages
  469--477, 2016.

\bibitem{luc2016semantic}
P.~Luc, C.~Couprie, S.~Chintala, and J.~Verbeek.
\newblock Semantic segmentation using adversarial networks.
\newblock {\em arXiv preprint arXiv:1611.08408}, 2016.

\bibitem{mao2017least}
X.~Mao, Q.~Li, H.~Xie, R.~Y. Lau, Z.~Wang, and S.~P. Smolley.
\newblock Least squares generative adversarial networks.
\newblock In {\em Computer Vision (ICCV), 2017 IEEE International Conference
  on}, pages 2813--2821. IEEE, 2017.

\bibitem{mathieu2016disentangling}
M.~F. Mathieu, J.~J. Zhao, J.~Zhao, A.~Ramesh, P.~Sprechmann, and Y.~LeCun.
\newblock Disentangling factors of variation in deep representation using
  adversarial training.
\newblock In {\em Advances in Neural Information Processing Systems}, pages
  5040--5048, 2016.

\bibitem{mcnamara2005exploring}
A.~McNamara et~al.
\newblock Exploring perceptual equivalence between real and simulated imagery.
\newblock In {\em Proceedings of the 2nd symposium on Applied Perception in
  Graphics and Visualization}, pages 123--128. ACM, 2005.

\bibitem{metz2016unrolled}
L.~Metz, B.~Poole, D.~Pfau, and J.~Sohl-Dickstein.
\newblock Unrolled generative adversarial networks.
\newblock {\em arXiv preprint arXiv:1611.02163}, 2016.

\bibitem{meyer1986experimental}
G.~W. Meyer, H.~E. Rushmeier, M.~F. Cohen, D.~P. Greenberg, and K.~E. Torrance.
\newblock An experimental evaluation of computer graphics imagery.
\newblock {\em ACM Transactions on Graphics (TOG)}, 5(1):30--50, 1986.

\bibitem{payne2005indoor}
A.~Payne and S.~Singh.
\newblock Indoor vs. outdoor scene classification in digital photographs.
\newblock {\em Pattern Recognition}, 38(10):1533--1545, 2005.

\bibitem{poole2016improved}
B.~Poole, A.~A. Alemi, J.~Sohl-Dickstein, and A.~Angelova.
\newblock Improved generator objectives for gans.
\newblock {\em arXiv preprint arXiv:1612.02780}, 2016.

\bibitem{portilla2000parametric}
J.~Portilla and E.~P. Simoncelli.
\newblock A parametric texture model based on joint statistics of complex
  wavelet coefficients.
\newblock {\em International journal of computer vision}, 40(1):49--70, 2000.

\bibitem{radford2015unsupervised}
A.~Radford, L.~Metz, and S.~Chintala.
\newblock Unsupervised representation learning with deep convolutional
  generative adversarial networks.
\newblock {\em arXiv preprint arXiv:1511.06434}, 2015.

\bibitem{reed2016generative}
S.~Reed, Z.~Akata, X.~Yan, L.~Logeswaran, B.~Schiele, and H.~Lee.
\newblock Generative adversarial text to image synthesis.
\newblock {\em arXiv preprint arXiv:1605.05396}, 2016.

\bibitem{salimans2016improved}
T.~Salimans, I.~Goodfellow, W.~Zaremba, V.~Cheung, A.~Radford, and X.~Chen.
\newblock Improved techniques for training gans.
\newblock In {\em Advances in Neural Information Processing Systems}, pages
  2234--2242, 2016.

\bibitem{shi2016real}
W.~Shi, J.~Caballero, F.~Husz{\'a}r, J.~Totz, A.~P. Aitken, R.~Bishop,
  D.~Rueckert, and Z.~Wang.
\newblock Real-time single image and video super-resolution using an efficient
  sub-pixel convolutional neural network.
\newblock In {\em Proceedings of the IEEE Conference on Computer Vision and
  Pattern Recognition}, pages 1874--1883, 2016.

\bibitem{torralba200880}
A.~Torralba, R.~Fergus, and W.~T. Freeman.
\newblock 80 million tiny images: A large data set for nonparametric object and
  scene recognition.
\newblock {\em IEEE transactions on pattern analysis and machine intelligence},
  30(11):1958--1970, 2008.

\bibitem{van2016pixel}
A.~Van~Oord, N.~Kalchbrenner, and K.~Kavukcuoglu.
\newblock Pixel recurrent neural networks.
\newblock In {\em International Conference on Machine Learning}, pages
  1747--1756, 2016.

\bibitem{wei2018improving}
X.~Wei, B.~Gong, Z.~Liu, W.~Lu, and L.~Wang.
\newblock Improving the improved training of wasserstein gans: A consistency
  term and its dual effect.
\newblock {\em arXiv preprint arXiv:1803.01541}, 2018.

\bibitem{winawer2008motion}
J.~Winawer, A.~C. Huk, and L.~Boroditsky.
\newblock A motion aftereffect from still photographs depicting motion.
\newblock {\em Psychological Science}, 19(3):276--283, 2008.

\bibitem{zhang2017deep}
M.~Zhang, K.~T. Ma, J.~H. Lim, Q.~Zhao, and J.~Feng.
\newblock Deep future gaze: Gaze anticipation on egocentric videos using
  adversarial networks.
\newblock In {\em IEEE Conference on Computer Vision and Pattern Recognition},
  pages 4372--4381, 2017.

\bibitem{zhao2016energy}
J.~Zhao, M.~Mathieu, and Y.~LeCun.
\newblock Energy-based generative adversarial network.
\newblock {\em arXiv preprint arXiv:1609.03126}, 2016.

\bibitem{zhu2016generative}
J.-Y. Zhu, P.~Kr{\"a}henb{\"u}hl, E.~Shechtman, and A.~A. Efros.
\newblock Generative visual manipulation on the natural image manifold.
\newblock In {\em European Conference on Computer Vision}, pages 597--613.
  Springer, 2016.

\bibitem{zhu2017unpaired}
J.-Y. Zhu, T.~Park, P.~Isola, and A.~A. Efros.
\newblock Unpaired image-to-image translation using cycle-consistent
  adversarial networks.
\newblock {\em arXiv preprint arXiv:1703.10593}, 2017.

\end{thebibliography}
}

\clearpage
\section{Supplementary Materials}
\subsection{Training Different Attribute Nets}
We implement attribute nets with three settings, VGG-16, ResNet50, and DenseNet169. In addition, we also conduct an experiment that substitutes the attribute net with random noise. The purpose is to check the effectiveness of different network architectures. We impose random noise to prove that only semantic vectors can improve the quality of generated images.

We train all the attribute nets on the annotated data. The dataset is randomly split into training and validation sets, each containing 500 images. All the images are resized to $224\times224$ pixels and normalized. The loss function is the mean squared error between predicted values and annotated values. We train the model for maximum 300 epochs. Training is stopped either the maximum epoch is reached or the loss plateaus on the validation set. All the models are optimized with the stochastic gradient descent method with a mini batch size 16.

\textbf{VGG-16.} The base model is a VGG net with 16 layers pretrained on ImageNet. The fully connected layer is replaced with a two-layer feed-forward neural network. The output dimension of each layer is 512 and 8 respectively. We use an initial learning rate of 0.02 and learning rate decay of 0.0001. The momentum is 0.9. The learning rate is halved for every 50 epochs. The size of a mini batch for each iteration is 8. The model is trained on an NVIDIA Titan Black GPU with 6 GB memory. The training takes about 12 hours to complete.

\textbf{ResNet-50.} The base model is a residual net with 50 layers pretrained on ImageNet. We change the output dimension of the last layer to 8. We use an initial learning rate of 0.01. The momentum is 0.9. The learning rate is halved for every 50 epochs. The size of a mini batch for each iteration is 8. The model is trained on an NVIDIA Titan Black GPU with 6 GB memory. The training takes about 10 hours to complete.

\textbf{DenseNet-169.} The base model is a dense net with 169 layers pretrained on ImageNet. We change the output dimension of the last layer to 8. We use an initial learning rate of 0.01. The momentum is 0.9. The learning rate is halved for every 50 epochs. The size of a mini batch for each iteration is 8. The model is trained on an NVIDIA Titan Black GPU with 6 GB memory. The training takes about 10 hours to complete.

\textbf{Random Noise.} Finally, we disable the attribute net and replace its output with a random noise vector $z\in\mathbb{R}^8$ drawn from $\mathcal{N}(\mathbf{0}, \mathbf{I})$.

\subsection{Training with Different Types of GANs}
We test the attribute net on three GAN variants, WGAN, DCGAN, and LSGAN. For a fair comparison, we use the original loss function and network architecture for each GAN. We train all the models on the tiny ImageNet dataset. We use the training set that consists of 1.28 million images to train the model. To monitor overfitting, we compute convergence curves of the discriminator’s value on both the training set and a test set that contains 50 thousand images. All the images have a resolution of $64\times64$ pixels. They are normalized without resizing before fed to the discriminator. We call the original GAN variant the vanilla model and the GAN variant with attribute net the modified model.

\textbf{WGAN.} The architectures of the generator and discriminator are the same as shown in Figure \ref{I}. The objective functions for training are defined as Equation (\ref{wgan_loss}) with gradient penalty term 10. We use the gradient penalty to regularize the generator and discriminator while keeping the attribute net fixed. We train the model for $200,000$ iterations. We use the Adam method with momentum terms $\beta_1=0$ and $\beta_2=0.99$ to optimize the model. The initial learning rate is 0.0001 and is kept constant during training. We update the discriminator once for every generator iteration. We use a mini batch size of 32 for each iteration. The model is trained on an NVIDIA Titan X GPU with 12GB memory. The training takes about 3 days for the vanilla model and 4 days for the modified model.

\textbf{DCGAN.} Let \texttt{k3n256s2} denote a $3\times3$ convolutional block with $256$ filters and stride $2$. \texttt{d256} denotes a $5\times5$ convolutional layer with 256 filters and stride 2. \texttt{fc4$\times$4$\times$512} denotes a fully connected layer with 4$\times$4$\times$512 filters and the output is reshaped to a $4\times4\times512$ tensor. The architecture of DCGAN is defined as following.

\begin{itemize}
	\item Generator: \texttt{fc4$\times$4$\times$512}, \texttt{d256}, \texttt{d128}, \texttt{d64}, \texttt{d3}, \texttt{tanh}.
	\item Discriminator: \texttt{k5n64s2}, \texttt{k5n128s2}, \texttt{k5n256s2}, \texttt{k5n512s2}, \texttt{fc1}
\end{itemize} The objective function to optimize is
\[
\min_G\max_DV(D,G)=\mathbb{E}_{p_{\text{data}}}[\log D(x)]+\mathbb{E}_{p_z}[\log(1-D(G(z)))].
\]
We use the Adam optimizer to train the model. Momentum terms are set to 0.5 and 0.999 respectively. We train the model for 200000 iterations. The initial learning rate is 0.0002 and is kept constant during training. We update the discriminator once for every generator iteration. We use a mini batch size of 32 for each iteration. The model is trained on an NVIDIA Titan X GPU with 12GB memory. The training takes about 2.5 days for the vanilla model and 3.5 days for the modified model.

\textbf{LSGAN.} LSGAN uses the same network architecture as DCGAN, but with different objective functions to optimize:
\[
\min_DV(D)=\frac{1}{2}\mathbb{E}_{p_{\text{data}}}[(D(x)-1)^2]+\frac{1}{2}\mathbb{E}_{p_z}[(D(G(z))+1)^2]
\]
and
\[
\min_GV(G)=\frac{1}{2}\mathbb{E}_{p_z}[(D(G(z)))^2].
\]
We use RMSProp to train the model. We train the model for 200000 iterations. The initial learning rate is 0.0001 and is kept constant during training. We update the discriminator once for every generator iteration. We use a mini batch size of 32 for each iteration. The model is trained on an NVIDIA Titan X GPU with 12GB memory. The training takes about 2.5 days for the vanilla model and 3 days for the modified model.

\subsection{Training on Different Datasets}
We also evaluate the attribute on the CIFAR-10 dataset as well. CIFAR-10 contains 60,000 images with a size of $32\times32$. We use the training set that contains 50,000 images to train the discriminator, and the remaining 10,000 for validation purpose. We evaluate WGAN with VGG as attribute net on the CIFAR-10 dataset. We train the model for 200000 iterations. We use gradient penalty to regularize norms of gradients. The regularization coefficient is 10. Initial learning rate is 0.0001 and is kept constant during training. We update the discriminator once for every generator iteration. We use a mini batch size of 32 for each iteration. The model is trained on an NVIDIA Titan X GPU with 12GB memory. The training takes about 2 days.

\begin{figure*}[htbp!]
	\centering
	\subfigure{
		\includegraphics[scale=0.3]{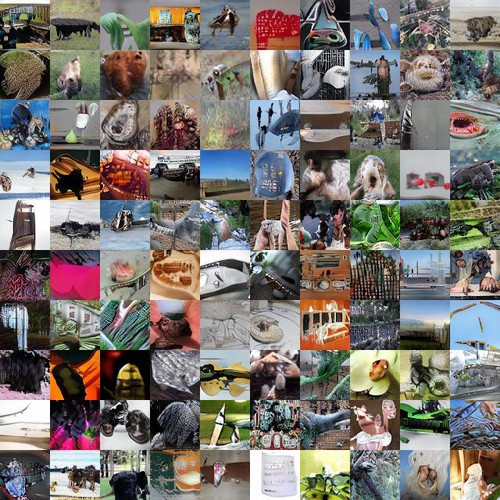}}
	\subfigure{
		\includegraphics[scale=0.3]{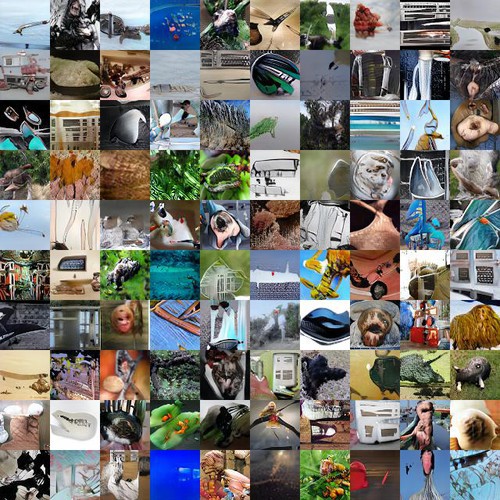}}
	\subfigure{
		\includegraphics[scale=0.3]{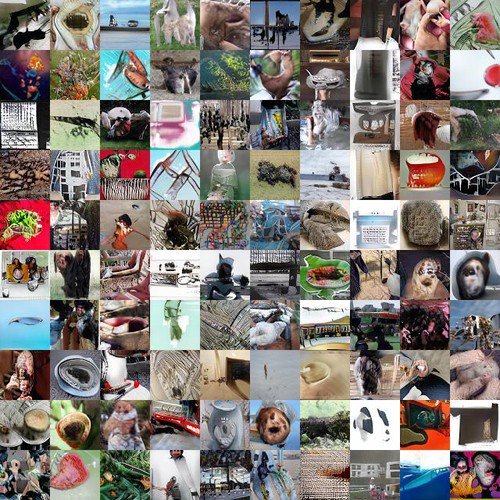}}
	\subfigure{
		\includegraphics[scale=0.3]{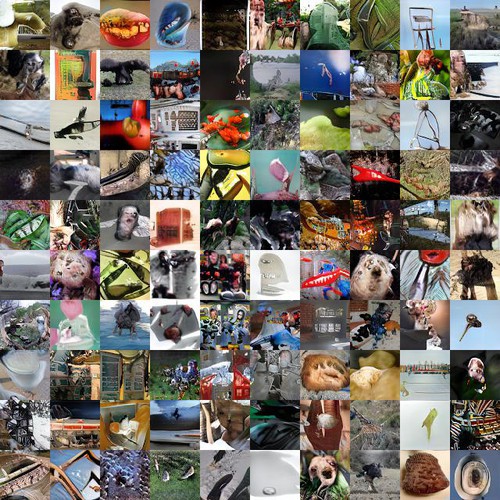}}
	\caption{More samples generated by our model (WGAN+VGG16).}\label{more1}
\end{figure*} 
\begin{figure*}[htbp!]
	\centering
	\subfigure{
		\includegraphics[scale=0.3]{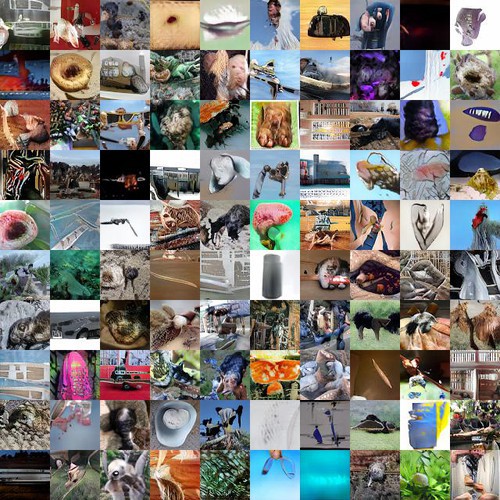}}
	\subfigure{
		\includegraphics[scale=0.3]{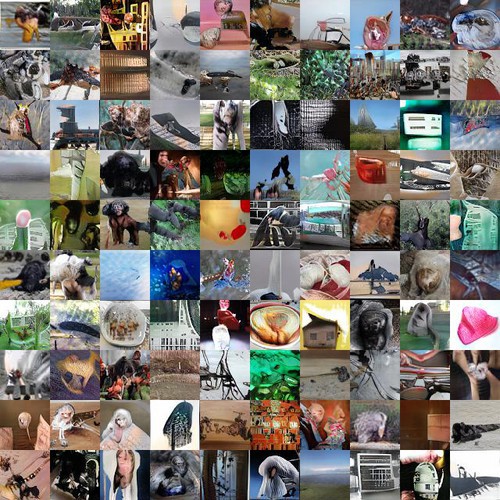}}
	\subfigure{
		\includegraphics[scale=0.3]{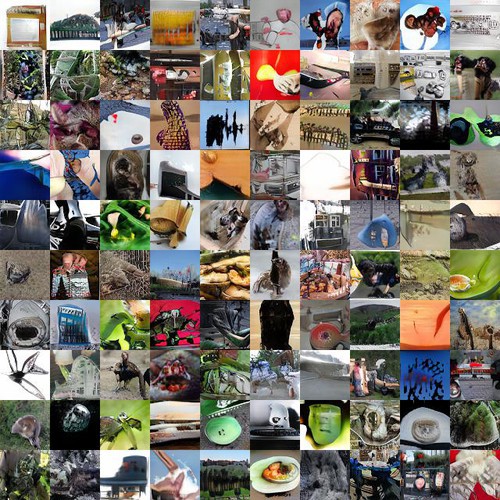}}
	\subfigure{
		\includegraphics[scale=0.3]{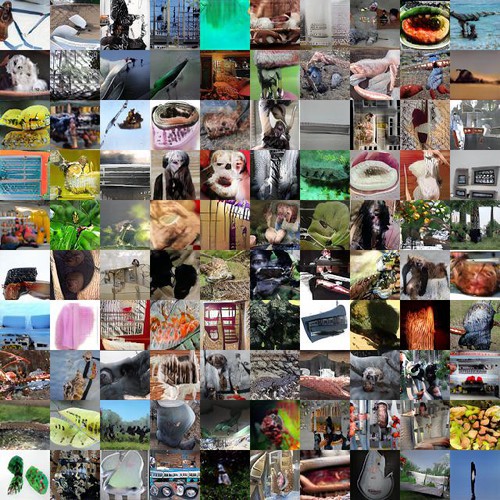}}
	\subfigure{
		\includegraphics[scale=0.3]{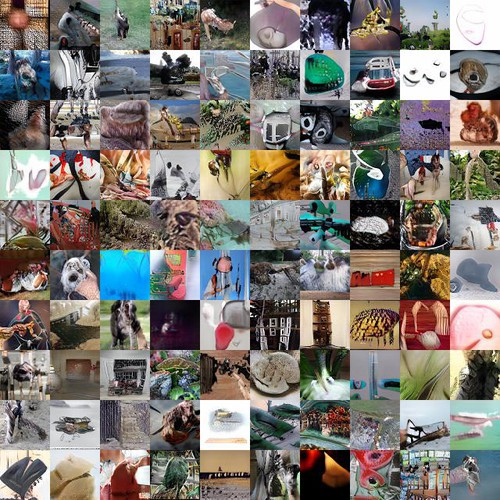}}
	\subfigure{
		\includegraphics[scale=0.3]{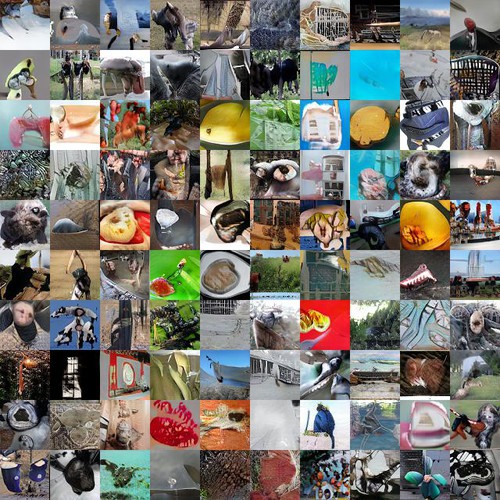}}
	\caption{More samples generated by our model (WGAN+VGG16).}\label{more2}
\end{figure*} 

\begin{figure*}[htbp!]
	\centering
	\subfigure[WGAN]{
		\includegraphics[scale=0.41]{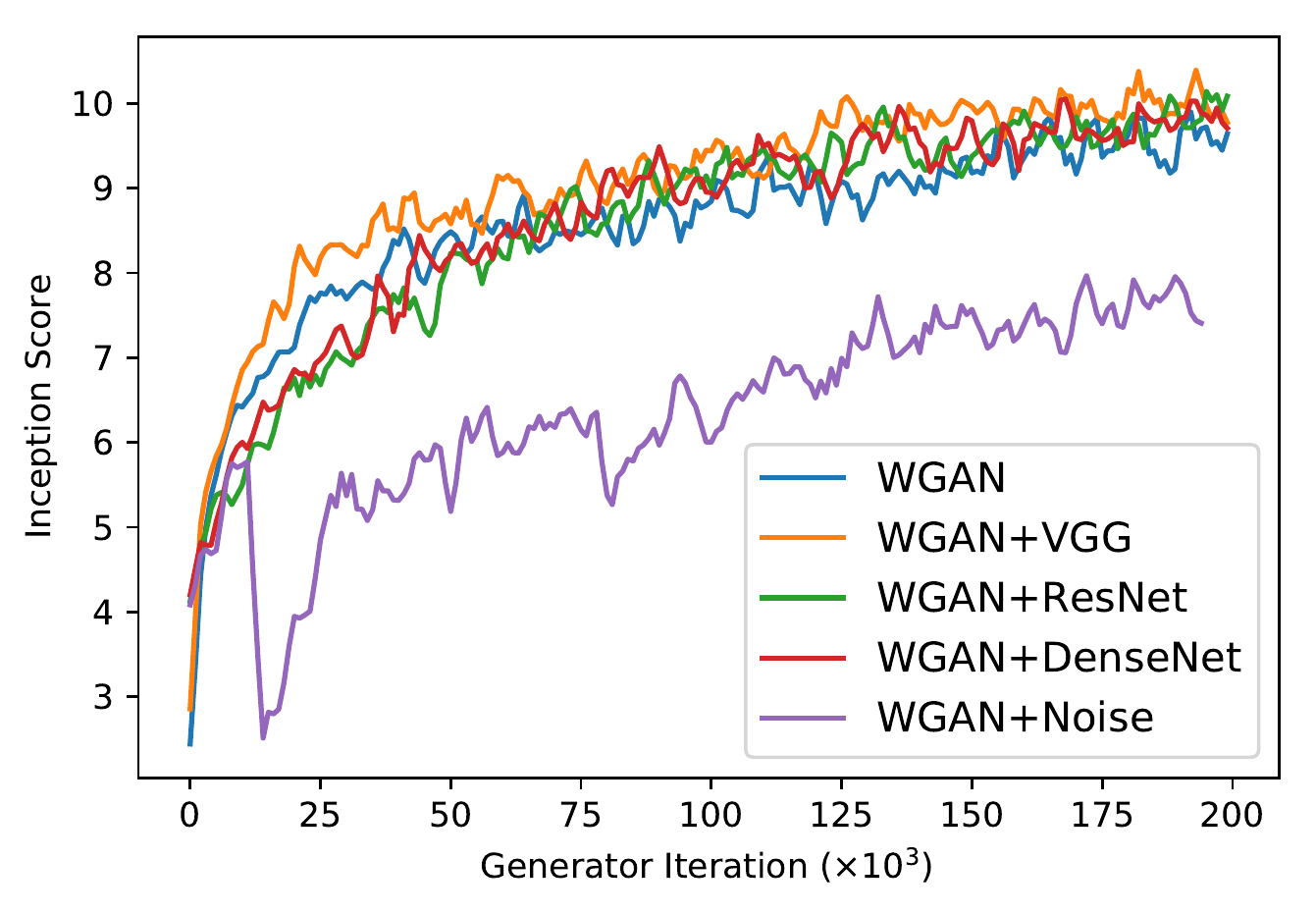}\label{wgan_ap}
	}    
	\subfigure[DCGAN]{
		\includegraphics[scale=0.41]{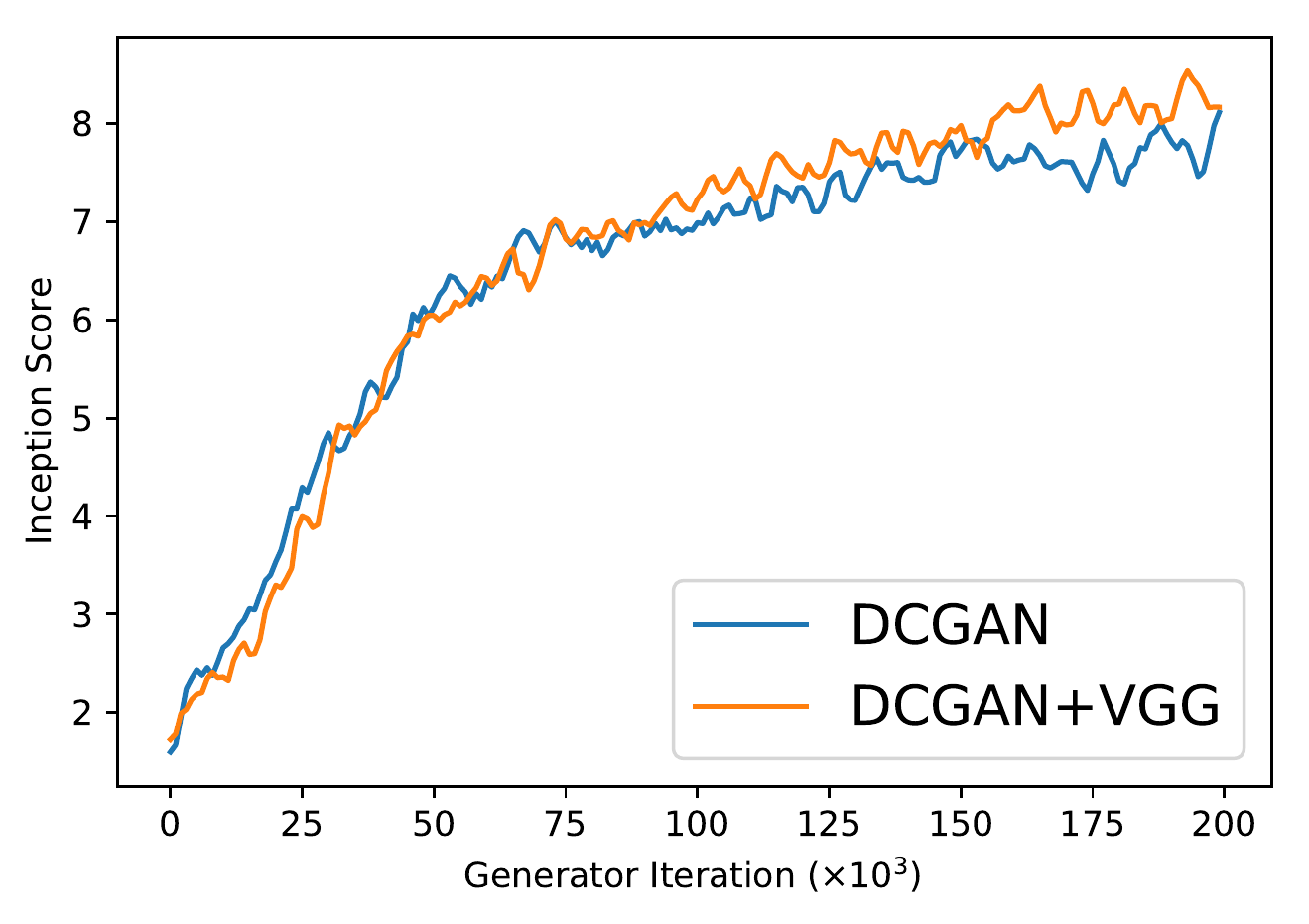}\label{dcgan_ap}
	}
	\subfigure[LSGAN]{
		\includegraphics[scale=0.41]{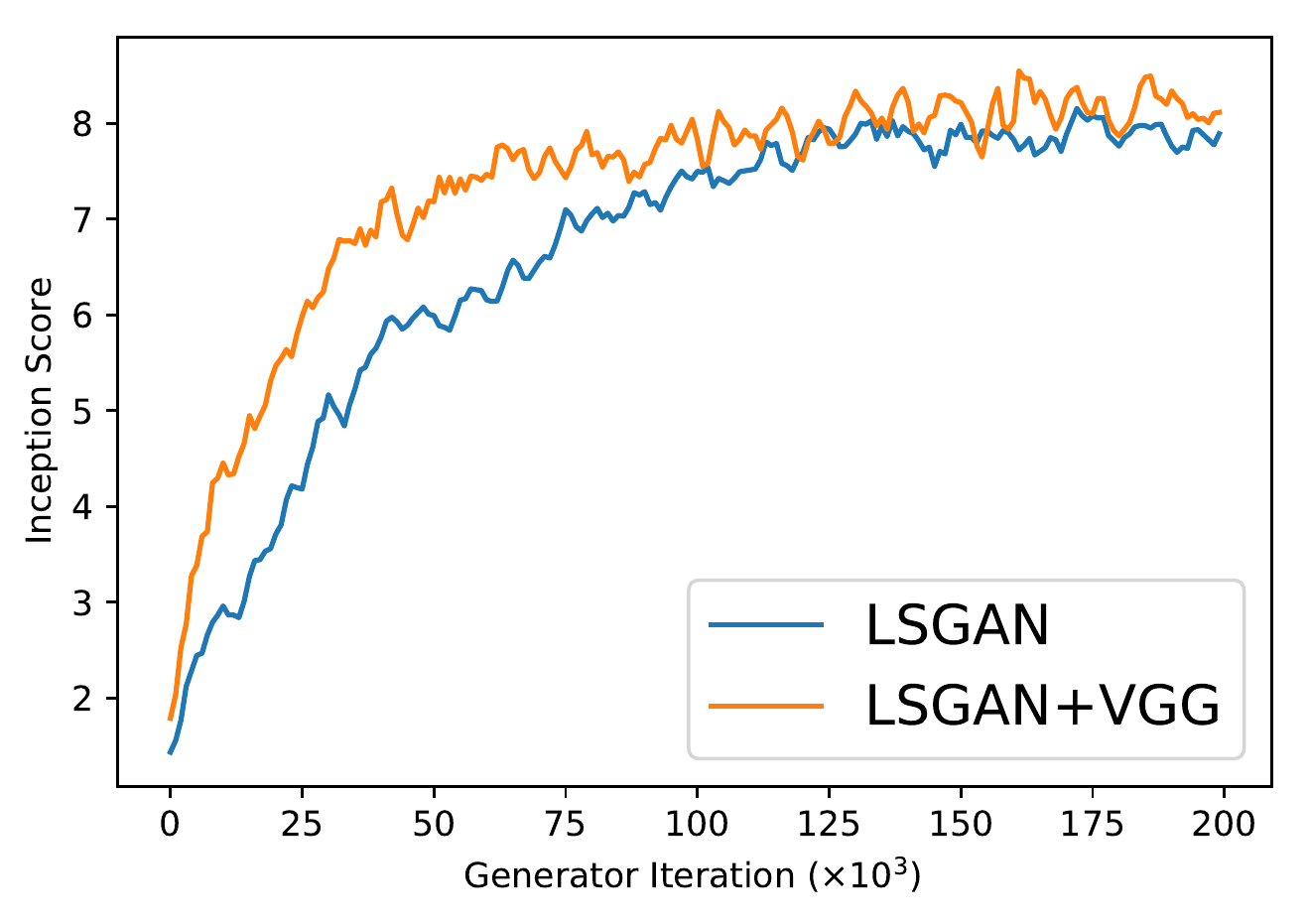}\label{lsgan_ap}
	}
	\caption{Training Curve. \ref{wgan_ap} shows that when attribute net is VGG the model converges fastest. When the attribute vector is replaced by a random vector, the inception score drops significantly. \ref{dcgan_ap} and \ref{lsgan_ap} show that adding an attribute net also improves the performance of other variants of GANs.}
	\label{training_curve}
\end{figure*} 
\end{document}